\documentclass[lettersize,conference]{IEEEtran}
\UseRawInputEncoding
\IEEEoverridecommandlockouts
\usepackage[letterpaper,top=54pt,bottom=54pt,left=54pt,right=54pt]{geometry}

\usepackage{times}
\usepackage{amsmath,amssymb,amsopn,amstext,amsfonts}
\usepackage{cancel}
\usepackage[space]{cite}
\usepackage{balance}
\usepackage{color}
\usepackage{mathtools}
\usepackage[ruled,vlined,linesnumbered]{algorithm2e}

\usepackage{bm}

\usepackage{diagbox}
\usepackage{float}
\usepackage{pifont}
\usepackage{amsmath}
\usepackage{multirow}
\usepackage{booktabs}
\usepackage{url}
\usepackage{svg}
\usepackage{threeparttable}
\usepackage{makecell}

\usepackage{array}
\usepackage{marvosym}
\usepackage{hyperref}           

\DeclareGraphicsExtensions{.png,.jpg,.eps,.pdf}
\DeclareGraphicsExtensions{.pdf}

\title{\LARGE \bf
\textmd{\textsf{FLIP}}: Real-Time and Resilient \textmd{\textsf{F}}ormation Planning for \textmd{\textsf{L}}arge-Scale D\textmd{\textsf{I}}stributed Swarms via \textmd{\textsf{P}}oint Cloud Registration
}

\author{Yuan Zhou, Guangtong Xu, Zhenyu Hou, Jialiang Hou, and Fei Gao
\thanks{This work was supported by the National Key R\&D Program of China under Grant No. 2023YFB4706600, the Zhejiang Provincial Science and Technology Plan Project under Grant No. 2024C01170 and the National Natural Science Foundation of China under Grant Nos. 62322314 \& 62203256. \emph{(Corresponding Author: Jialiang Hou; Fei Gao.)}}
\thanks{Yuan Zhou, Jialiang Hou, and Fei Gao are with the Institute of Cyber-Systems and Control, College of Control Science and Engineering, Zhejiang University, Hangzhou 310027, China, and also with the Huzhou Institute, Zhejiang University, Huzhou 313000, China. (e-mail: \{y2zhou, fgaoaa\}@zju.edu.cn, jlhou8@gmail.com). Zhenyu Hou and Guangtong Xu is with the Huzhou Institute, Zhejiang University, Huzhou 313000, China. (e-mail: xiagelearn@gmail.com, guangtong\_xu@163.com)}
}

\begin{document}

\maketitle

\begin{abstract}
Traditional large-scale formation planning either oversimplify the formation representation which leads to poor performance, or they employ complete collaborative relationships, which results in excessive computational load.
To achieve high-performance and large-scale formation planning, we transform the Optimal Formation Position Sequence \cite{c1} (OFPS) calculation problem into a spatiotemporal Point Cloud Registration (PCR) problem. Each agent derives its OFPS by distributively computing the matching result between current positions and the desired formation positions of all other agents.
Then each agent optimizes the cooperative formation trajectory by using OFPS. We leverage the PCR method with outlier rejection to rapidly perform large-scale formation position registration. This prevents suboptimal trajectories and failed agents from propagating through the cooperative network and affecting more agents. Consequently, we uniformly achieve resilient, efficient, and distributed trajectory planning for large-scale swarms. The effectiveness and the superiority of the proposed method are demonstrated through large-scale simulations of 120-drone formation, and rigorous benchmarking against state-of-the-art (SOTA) methods.

\end{abstract}


\section{Introduction}
\label{sec:introduction}
Formation planning is a fundamental capability for swarm robots. In recent years, research focus on applying large-scale formations in complex environments to extend the applications of swarm robots \cite{c1}, \cite{c18}, \cite{c16}. However, for large-scale formation planning \cite{1000agents}, \cite{1mswarm}, \cite{vrb}, the increase in scale introduces more constraints, which not only consumes computational resources but also makes the system more prone to become trapped in local optima. Furthermore, these local solutions, along with the frequently occurring agent failure states, can propagate through the collaborative network, affecting more agents and leading to greater formation errors. For large-scale formation, this can severely degrade formation performance and may even result in the ultimate failure of the formation.

\begin{figure}[!t]
    \centering
    \includegraphics[width=3.5in]{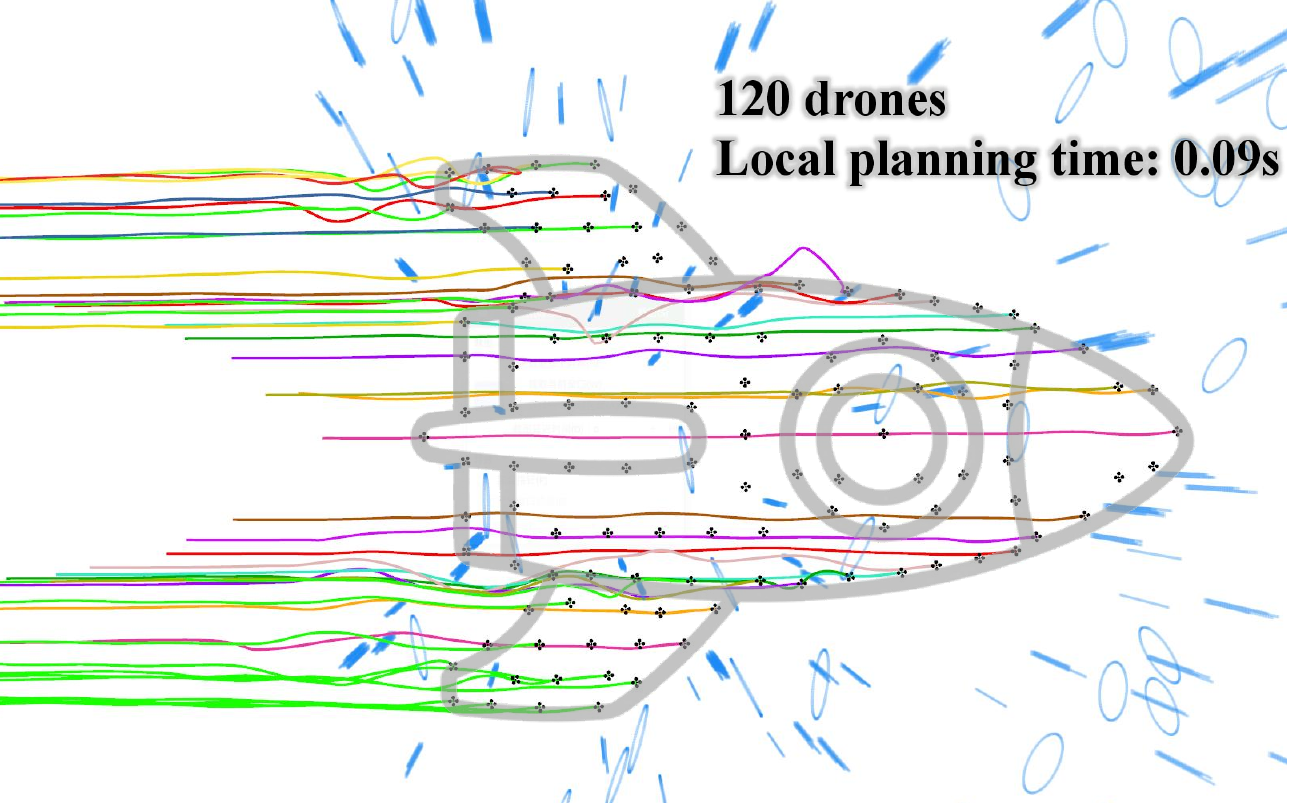}
    \vspace{-0.5cm}
    \caption{
    Simulation results of PCR-based efficient formation planning, maintaining a rocket-shaped of 120 drones while avoiding obstacles.
    }
    \vspace{-0.3cm}
    \label{fig1}
    \vspace{-0.4cm}
\end{figure}
To achieve large-scale formation planning, some studies characterize the overall desired formation using a small set of parameters. Examples include employing polyhedral structures \cite{c16}, probability distributions \cite{1mswarm}, and artificial potential fields \cite{zhaoswarm} to represent formations and guide coordination. These methods decouple collaborative connections between most agents, meaning that poor performance by some agents does not adversely affect others. However, this oversimplified formation representation and the simplistic collaborative network lead to a significant degradation in performance. In obstacle environments, this often results in substantial shape deformation or slow convergence speed for the formation. 
To ensure formation performance, some works meticulously consider the impact of each agent's motion on the overall formation maintenance \cite{c1}, \cite{peng}, \cite{c15}. The collaborative network between agents approximates a complete graph. As the scale increases, those methods introduce excessive collaborative constraints, which not only elevate the computational load, but also make the system prone to converging to locally optimal solutions. Furthermore, due to the lack of a mechanism for handling abnormal agent movements, suboptimal solutions or aberrant motions of some agents can easily propagate through the collaborative network \cite{zhou2025re}, affecting more agents and thus compromising long-term formation planning. To balance formation performance and computational load, some studies adopt multi-leader approaches or rigid sparse graphs \cite{c25}, \cite{c22} to construct sparse collaborative networks, even incorporating mechanisms for handling abnormal agents \cite{zhou2025re}. However, as the scale increases over 100, sparse collaborative networks fail to guaranty formation performance. Moreover, \cite{c1}, \cite{zhou2025re} uses the normalized Laplacian matrix for elongated formations, which results in major-axis gradients exceeding minor-axis ones and affects performance.

\begin{figure}[!t]
    \centering
    \includegraphics[width=3.6in]{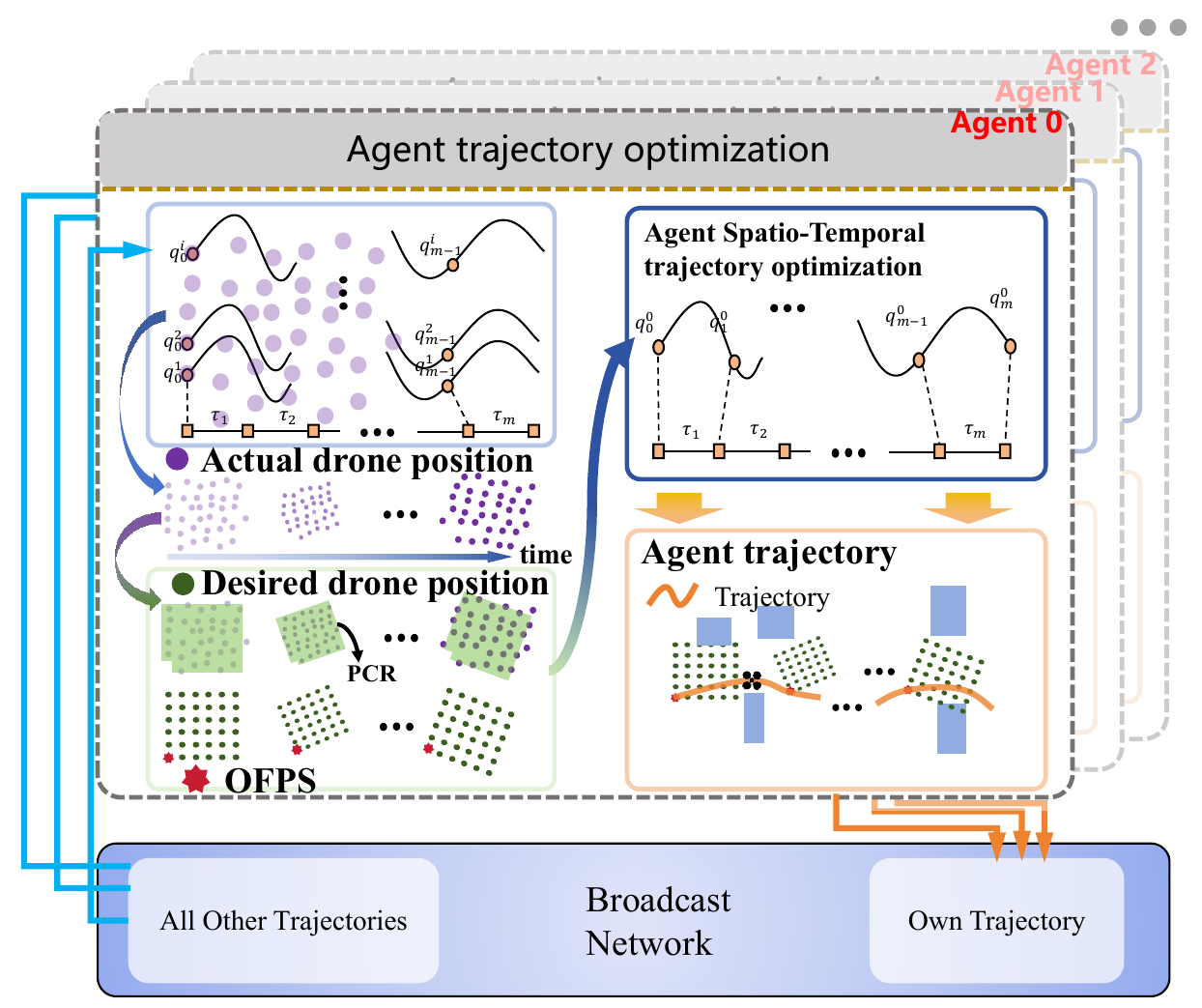}
    \vspace{-0.5cm}
    \caption{
    Our formation planning algorithm flowchart. The system comprises the OFPS calculation based on PCR, trajectory optimization, and wireless communication. Each agent receives the trajectories of all other agents and derives a series of position distributions for them over a future time horizon based on the sampling time. These position points are treated as point clouds and subjected to PCR with the desired formation positions to obtain a corresponding series of matching parameters. Based on these parameters and its own desired formation position, the agent calculates the OFPS. Finally, the agent treats the OFPS as constraints to perform formation trajectory optimization and broadcast local trajectory to all others.
    }
    \vspace{-0.3cm}
    \label{framework}
    \vspace{-0.2cm}
\end{figure}

Unlike the above methods, to achieve efficient and resilient large-scale formation planning, we treat swarms in formation planning as points in the PCR problem \cite{maxcline}. By transforming the OFPS calculation problem \cite{c1} into a series of PCR problems for the spatiotemporal positions of the swarms and applying efficient PCR algorithms with outlier rejection \cite{random}, \cite{pcm}, \cite{teaser} to compute OFPS, we meticulously coordinate all agents to ensure formation performance and inherently eliminate the effects of abnormal states and suboptimal trajectories of some agents. Specifically, as illustrated in Fig. \ref{framework}, we adopt a distributed asynchronous swarm planning framework. Each agent obtains the positional distribution of other agents over a horizon based on broadcast trajectories. To prevent the problem where the normalized Laplacian matrix struggles to handle elongated formations, we do not adopt a graph-based formation representation but use a point cloud for representation. At each time step, the positions of other agents are treated as the point cloud, while the desired formation is regarded as the target point cloud. By applying the mature PCR algorithm, the optimal transformation parameters between the two point clouds are computed. In this way, we can calculate the OFPS by the optimal transformation parameters and the initial desired formation position, then the OFPS is used as a formation constraint for single-agent trajectory optimization of formation coordination. In this paper, we employ RANSAC \cite{random} to eliminate abnormal agent positions while maintaining computational efficiency. We demonstrate effectiveness and superiority through simulations and benchmarks that our method can handle real-time formation planning and can achieve formation planning for 120 agents. Our method maintains formation for normal agents even with nearly 10\% anomalous agents. Our contributions are as follows:
\begin{enumerate}
    \item We propose a large-scale formation planning algorithm based on PCR. Through the application of an efficient PCR algorithm to compute the optimal formation position sequence, our method achieves large-scale formation real-time planning.
    \item We achieve resilient formation planning by employing a PCR algorithm with outlier rejection. This approach eliminates the influence of suboptimal and abnormal trajectories on other agents during distributed trajectory optimization.
    \item We validate the effectiveness and advantages of our algorithm through extensive experiments and benchmarks, and we also open-source our code.
\end{enumerate}

\section{Related Work}
\label{sec:related_work}
We review existing research on large-scale formation navigation in complex environments. Furthermore, we extend the investigation to the development of resilient swarms, particularly strategies that account for potential agent failures.
\subsection{Lager-scale Formation in Complex Environments}
To achieve large-scale formations with limited computational resources, some research tends to simplify the representation of formation constraints while also simplifying motion planning algorithms, such as the artificial potential field (APF) \cite{1000agents}, \cite{zhaoswarm}, the probability distribution \cite{1mswarm}, \cite{markov} and polyhedral virtual structures \cite{c16}, \cite{vrb}. However, most of these methods do not account for complex obstacle environments, and their convergence speed to the desired formation in the presence of disturbances is relatively slow, making them difficult to achieve real-time formation planning.

Another category of methods meticulously accounts for the impact of each agent's motion on maintaining the overall desired formation. Some approaches employ centralized planning to generate trajectories for all agents \cite{c15}, \cite{peng}, while others adopt distributed formation planning based on the motion states of other agents \cite{c1}, \cite{c18}. However, these methods either consume excessive computational resources due to the proliferation of optimization variables or introduce complex nonlinear formation constraints. The sharply escalating computational complexity under fully connected collaborative relationships significantly restricts the scalability of the swarm scale. Reinforcement learning is applied successfully to maintain formations in obstacle environments \cite{learning2}, \cite{wuyi_formation}. However, these methods struggle with large-scale coordination in complex environments.
To balance formation performance with computational efficiency, some approaches adopt sparse collaborative topologies for formation coordination \cite{zhou2025re}, thereby reducing the complexity of constraints in the planning process. However, these methods still struggle to guarantee satisfactory formation performance in over 100 agents due to limitations in computational resources.

\subsection{Resilient Swarm System and Outliers Rejection}
Considering sub-optimally coordinated agents and anomalous agent states and to prevent these states from propagating through the coordination network and affecting other agents, various methods have been employed in numerous studies to mitigate their impact. Some studies design adaptive control laws for anomalous agent states to achieve formation control \cite{relient3}, \cite{relient4}. Some also adopt the approach of utilizing robust communication graphs for resilient formation control \cite{resilient1}, \cite{resilient2}. Those methods rely on assumptions about the model of agent anomalies and are not applied in complex environments. Zhou et al. \cite{zhou2025re} treat anomalous agents as outliers in formation planning and eliminate anomalous agents by computing the maximum clique. However, their approach performs anomaly removal in a single step before trajectory optimization, which struggles to handle suboptimal agent trajectories. Additionally, the method can only cope with a small number of anomalous individuals, resulting in limited robustness. In this paper, we transform the problem of OFPS in formation trajectory optimization into a series of PCR problems. The positions of these sub-optimally coordinated agents and the states of anomalous agents are likewise treated as outliers. There are many methods used to solve the problem of outliers rejection \cite{random}, \cite{pcm}, \cite{teaser}. We adopt RANSAC as the core algorithm for PCR due to its efficiency and the availability of open-source implementations.

\section{LARGE-SCALE FORMATION PLANNING BASED PCR}

\subsection{Formation Coordination and PCR}
\label{frond_end0}
In this section, we discuss the relationship between formation coordination and PCR, illustrating the rationality of PCR-based algorithm for formation planning. The objective of formation coordination is to make the relative positions of the swarm as close as possible to the desired formation. Inspired by \cite{c1}, \cite{zhou2025re}, \cite{how_formation}, we define the objective function for formation coordination at a certain moment:
\begin{align}
\begin{split}
&\hat e_{dist}= \mathop{\rm min}_{\mathbf R, \mathbf t, s} \sum_{i=1}^{N_{a}} || p_i^{des} - (s \mathbf R p_i^{cur}+\mathbf t)||_2, \label{eq1} \\
\end{split}
\end{align}
where $N_{a}$ is the number of all agents, $p_i^{des}$ and $p_i^{cur}$ represent the desired position and the current position of the $i$th robot in formation, respectively. A Sim(3) transformation consists of a rotation $\mathbf R \in SO(3)$, a translation $\mathbf t \in \mathbb R^3$, and a scale factor $s \in \mathbb R$. Essentially, such an objective function serves as the error calculation formula for a point cloud registration problem \cite{teaser}. 

Our framework takes into account the impact of all agents on overall formation coordination. We adopt a distributed and asynchronous swarm planning framework \cite{c1}, \cite{c18}, whereby agents optimize their own trajectories based on broadcasted trajectories received from other agents to achieve coordinated motion. During the optimization process, the trajectories of other agents are treated as fixed, with only the agent's own trajectory (i.e. position) being optimized. Within this framework, the solution for the optimal formation position (OFP) for agent $i$ is formulated as: 
\begin{align}
\begin{split}
& \mathop{\rm min}_{p_i} \mathcal F(p_i, \mathcal C), \label{eq2} \\
\end{split}
\end{align}
\begin{align}
\begin{split}
&\mathcal C  = \{ p_1, \dots, p_{i-1}, p_{i+1}, \dots, p_{N_a} \}, \label{const} \\
\end{split}
\end{align}
where $\mathcal F(\cdot)$ denotes a function that calculates the error relative to the desired formation. And all the elements in $\mathcal C$ are fixed values. According to the definition of the formation coordination objective in Eq. \ref{eq1}, we can see that regardless of the method used to characterize the desired formation shape and compute the optimal formation positions, our goal is to find the same optimal point $p^*_i$ in space such that:
\begin{equation}
\begin{aligned}
\mathop{\rm min}_{p_i} \mathcal F_{PCR}(p_i, \mathcal C), \label{eq3}
\end{aligned}
\end{equation}
\begin{equation}
\begin{aligned}
\mathcal F_{PCR}(p_i,\mathcal C)= 
\mathop{\rm min}_{\mathbf{R}, \mathbf{t}, s} 
\biggl( \sum_{j \in {N_a} \setminus \{i\}}^{N_a} \| p_j^{des} - (s \mathbf{R} p_j^{cur}+\mathbf{t}) \|_2 
\\
+ \| p_i^{des} - (s \mathbf{R} p_i+\mathbf{t}) \|_2 \biggr). \label{eq4}
\end{aligned}
\end{equation}
Assuming the number of optimization iterations for Eq. \ref{eq3} is $n_{f}$ and the number of optimization iterations for the PCR problem in Eq. \ref{eq4} is $m_f$, the total iteration times is $n_{f}m_{f}$. In the next section, we will discuss how to solve Eq. \ref{eq3} efficiently, and demonstrate how to solve the PCR problem once to approximate $p^*_i$.

\subsection{Efficient OFP Approximate Solution}
\label{frond_end00}
In Sect. \ref{frond_end0}, we observe that the formation position optimization problem in Eq. \ref{eq3} is equivalent to a PCR problem. However, the difficulty of iteratively solving for the optimal position $p^*_i$ increases as the number of agents $N_a$ grows in Eq. \ref{eq4}.

An approach to reduce the computational burden involves selecting $M$ agents from the total ${N_a}$ and solving problem:
\begin{equation}
\begin{aligned}
\mathop{\rm min}_{p_i} \mathcal F_{PCR}(p_i, \underbrace{\{ \dots, p_{j}, \dots \}}_{M}). \label{eq5}
\end{aligned}
\end{equation}
It is evident that Eq. \ref{eq4} \& Eq. \ref{eq5} still represent the point cloud registration problem. Since the positions of other agents are fixed values in the optimization problem and simultaneously constitute the majority of terms in this PCR, the optimization solution for $\{ \mathbf{R}, \mathbf{t}, s\} $ is primarily influenced by $\sum_{j \in {N_a} \setminus \{i\}}^{N_a} \| p_j^{des} - (s \mathbf{R} p_j^{cur}+\mathbf{t}) \|_2$. Assuming that the optimal solution for Eq. \ref{eq4} is $\{ \mathbf{R}^*, \mathbf{t}^*, s^*\} $. When the optimal solution $\{ \mathbf{R}^+, \mathbf{t}^+, s^+\} $ of the problem Eq. \ref{eq5} constructed by selecting $M$ points is close to $\{ \mathbf{R}^*, \mathbf{t}^*, s^*\} $, it can obtain an approximate solution $p^+_i$ to trade off precision for computational efficiency. 

The key to the problem is selecting $M$ points that can ensure that $p^+_i$ and $p^*_i$ are as close as possible.
Zhou et al. \cite{zhou2025re} demonstrate that selecting agents located on the outer contour of the formation shape ensures better formation performance. As shown in the proof in the Appendix \ref{sec:appendix} , it can be readily known that points farther from the centroid (i.e., those on the outer contour of the point cloud) have greater influence on the overall outcome. Since Eq. \ref{eq4} is a PCR problem, the conclusions in the appendix also apply to it. It has been experimentally verified that a slight sacrifice of this error can still ensure satisfactory formation performance in \cite{zhou2025re}. However, as the scale increases, a small number of $M$ points becomes inadequate to represent the overall formation, leading to a growing discrepancy between $\{ \mathbf{R}^*, \mathbf{t}^*, s^*\} $ and $\{ \mathbf{R}^+, \mathbf{t}^+, s^+\} $, and thus failing to ensure satisfactory formation performance. 

The above method demonstrate that formation can be maintained without position information from all agents. However, as the number of agents continues to increase, sparse graph based method \cite{zhou2025re} remain constrained by the increasing number of constraints. In this paper, we do not use position information from all agents for coordination. However, compared to Eq. \ref{eq5}, we only exclude the point $p_i$ in Eq. \ref{eq4} to approximate the solution of $p^*_i$, thereby theoretically guaranteeing formation performance close to the complete graph formation.
Analysis of Eq. \ref{eq3} \& Eq. \ref{eq4} reveals that obtaining a precise solution for $p^*_i$ requires multiple optimization iterations. In practice, for large-scale formation planning, the position of a single point $p_i$ in Eq. \ref{eq4} has a minor impact on the solution of $\{ \mathbf{R}^*, \mathbf{t}^*, s^*\} $. Therefore, we choose to sacrifice precision by ignoring $\| p_i^{des} - (s \mathbf{R} p_i+\mathbf{t}) \|_2$ in Eq. \ref{eq4}. Therefore, Eq. \ref{eq4} is relaxed to:
\begin{equation}
\begin{aligned}
\mathop{\rm min}_{\mathbf{R}, \mathbf{t}, s} 
\sum_{j \in {N_a} \setminus \{i\}}^{N_a} \| p_j^{des} - (s \mathbf{R} p_j^{cur}+\mathbf{t}) \|_2 , \label{eq6}
\end{aligned}
\end{equation}
\begin{equation}
\begin{aligned}
p^\#_{i} = s^\# \mathbf{R}^\# p_i^{des}+\mathbf{t}^\#, \label{eq7}
\end{aligned}
\end{equation}
where $\{ \mathbf{R}^\#, \mathbf{t}^\#, s^\#\} $ is the optimal solution of Eq. \ref{eq6}. $p^\#_{i}$ is the 
corresponding optimal formation position. Since we have considered all other agents positions to solve Eq. \ref{eq6}, the error of $p^*_i$ and $p^\#_i$ are sufficiently small in large-scale formation.


\subsection{OFPS Calculation via PCR}
\label{frond_end}
In Sect. \ref{frond_end00}, we transform the OFP generation into a PCR problem. In this section, we collect trajectories from all other agents and compute a series of OFP over a future horizon (i.e. OFPS) using the well-established PCR algorithm. Assume that there is a sequence of time stamps $\{0,\dots,m,\dots, M_c\}$ for the future period. Following the approach in \cite{c1}, we compute the optimal formation position sequence $\mathbf{p}^\#_{i,f} = \{ {p^\#}_{i,f}(0),\dots,{p^\#}_{i,f}(m),\dots, {p^\#}_{i,f}(M_c)\}$ based on the trajectories of all other agents. 

We can first obtain positions sequence of other agents except for agent $i$ based on their broadcasted trajectories:
\begin{equation}
\begin{aligned}
\{\mathbf{P}_{i,all}(0),\dots,\mathbf{P}_{i,all}(m),\dots, \mathbf{P}_{i,all}(M_c)\}, \label{eq7}
\end{aligned}
\end{equation}
where $\mathbf{P}_{i,all}(m)$ represents the positions of all other communicating agents at time $m$:
\begin{equation}
\begin{aligned}
\mathbf{P}_{i,all}(m) = \{\dots,{p}_{j}(m),\dots, \}, j \in {N_a} \setminus \{i\}. \label{eq8}
\end{aligned}
\end{equation}
Then, we use the RANSAC algorithm \cite{random} for PCR for each $\mathbf{P}_{i,all}(m)$ and we can obtain the optimal solution $\{ \mathbf{R_m}^\#, \mathbf{t_m}^\#, s_m^\#\} $:
\begin{equation}
\begin{aligned}
\mathop{\rm min}_{\mathbf{R_m}, \mathbf{t_m}, s_m} 
\sum_{j \in {N_a} \setminus \{i\}}^{N_a} \| p_j^{des} - (s_m \cdot \mathbf{R_m} \cdot {p}_{j}(m)+\mathbf{t_m}) \|_2 . \label{eq9}
\end{aligned}
\end{equation}
The optimal formation position of agent $i$ on time stamp $m$, which is used as the formation constraint in the following agent trajectory optimization, is obtained by:
\begin{equation}
\begin{aligned}
{p^\#}_{i,f}(m) = s_m^\# \cdot \mathbf{R}_m^\# \cdot p_i^{des}+\mathbf{t}_m^\#. \label{eq10}
\end{aligned}
\end{equation}

\subsection{Trajectory Representation}
\label{traj_represent}
In this paper, we adopt $\mathfrak{T}_{\textbf{MINCO}}$ \cite{c32} as the trajectory representation. This class defines a polynomial trajectory with minimal control effort, characterized by:

\begin{align}
\begin{split}
\mathfrak{T}_{\textbf{MINCO}} = \{\bm \sigma(t):[0,T_\Sigma] \rightarrow \mathbb{R}^m | \textbf{c} = \mathcal{M}(\textbf{q},\textbf{T}), \\
\textbf{q} \in \mathbb{R}^{m(M-1)}, \textbf{T} \in \mathbb{R}^M_{>0}\},
\end{split}
\end{align}

where $p(t)$ represents an $m$-dimensional $M$-piece polynomial trajectory of degree $N = 2s-1$ and $s$ denotes the order of the relevant integrator chain. For the agent, $\bm \sigma(t)={p}^T(t)$. The polynomial coefficient $\textbf{c} = (\textbf{c}^T_1, ..., \textbf{c}^T_M)^T \in \mathbb{R}^{2Ms \times m}$ is derived from $\mathcal{M}(\textbf{q},\textbf{T})$, where $\textbf{q} = (\textbf{q}_1, ..., \textbf{q}_{M-1})$ represent the waypoints, $\textbf{T} = (T_1, T_2, ..., T_M)^T$ represents the time allocated for each trajectory segment. $T_\Sigma = \sum_{i=1}^M T_i$ signifies the total time span. Each $m$-dimensional $M$-piece trajectory is defined as follows:
\begin{align}
\bm \sigma(t) = \bm \sigma_i(t - t_{i-1}) \quad \text{for all } t \in [t_{i-1}, t_i),
\end{align}
we use a 5 order polynomial, and the $i^{th}$ piece trajectory is
\begin{align}
\bm \sigma_i(t) = \textbf{c}^T_i \beta(t) \quad \text{for all } t \in [0, T_i),
\end{align}
where $\textbf{c}_i \in \mathbb{R}^{(N+1) \times m}$ represents the coefficient matrix, $\beta(t) = [1, t, ..., t^N]^T$ is the natural basis and $T_i = t_i - t_{i-1}$ is the time allocated for the $i^{th}$ piece. The uniqueness of $\mathfrak{T}_{\textbf{MINCO}}$ is determined by $(\textbf{q}, \textbf{T})$. Through parameter mapping$\textbf{c} = \mathcal{M}(\textbf{q}, \textbf{T})$, trajectory representations $(\textbf{c}, \textbf{T})$ can be converted to $(\textbf{q}, \textbf{T})$. This conversion allows the expression of arbitrary second-order continuous cost functions $J(\mathcal{M}(\textbf{q}, \textbf{T}), \textbf{T})$ in the form $H(\textbf{q}, \textbf{T})$.

For enforcing time-continuous constraints including obstacle avoidance, dynamic viability, and terrain traversal cost, each trajectory piece is uniformly sampled in $\kappa_i$ temporal instances defined by $\tilde{p}_{i,j} = p_i((j / \kappa_i) \cdot T_i)$, for $j = 0, 1, ..., \kappa_i - 1$.

\subsection{Trajectory Optimization Constrained by OFPS}
\label{back_end}
After obtaining the optimal formation position sequence in Sect. \ref{frond_end}, we incorporate the position sequence as the formation constraint. Simultaneously, considering the agent dynamic constraint, collision avoidance, and objectives for optimal time and energy, we formulate agent trajectory planning as an optimization problem:

\begin{align}
& \min_{\textbf{q},\textbf{T}} \quad \int^{t_M}_{t_0} {\| p^{(s)}(t) \|}^{2}dt\ + \ \rho \cdot{T_\Sigma} , \\
& \text{s.t.} \quad p(t) = \mathcal{M}_{\textbf{q},\textbf{T}} \quad \forall t \in [t_0,t_M], \\
& \quad \quad \mathbf{p}^{[s-1]}(0) = \bar{\mathbf{p}}_0,\\
& \quad \quad \mathbf{p}^{[s-1]}(t_M) = \bar{\mathbf{p}}_f ,\\
& \quad \quad \mathcal{H}(p(t), ...,p^{(s)}(t)) \preceq 0 \quad \forall t \in [t_0,t_M] .
\end{align}

The trajectory $p(t)$ is defined by the optimization variables $\{ \textbf{q},\textbf{T}\}$ and the temporal regularization coefficient $\rho$. The state vector $\mathbf{p}^{[s-1]}(t) = (p(t)^T,\dot p(t)^T,...,p^{[s-1]}(t)^T)^T\in \mathbb R^{ms}$ encodes the $[s-1]^{th}$ order derivatives of a chained dynamical system governed by $s$-integrator dynamics. The initial state is $\bar {\mathbf{p}}_0$ and the terminal state is $\bar {\mathbf{p}}_f$. 

The continuum constraint set $\mathcal{H}$ encompasses dynamic feasibility $\mathcal{J}_d$, obstacle clearance $\mathcal{J}_o$, and swarm reciprocal avoidance $\mathcal{J}_s$. Those constraints are modeled by using the same method in \cite{c1}, and can refer to \cite{c1} for details. For formation coordination $\mathcal{J}_f$, we derive the desired spatiotemporal formation position sequence $\mathbf{p}^\#_{i,f} = \{ {p^\#}_{i,f}(0),\dots,{p^\#}_{i,f}(j),\dots, {p^\#}_{i,f}(M_c)\}$ over a future time period with a sequence of time stamps $\{0,\dots,j,\dots, M_c\}$. This establishes the formation constraint for the agent $i$ as follow
\begin{align}
\mathcal{J}_f \implies \sum_{j=0}^{M_c} \| p(j) - {p^\#}_{i,f}(j)\|^2.
\end{align}

The equality constraints Eq. \ref{eq1} \& Eq. \ref{eq3} are eliminated using MINCO variables, enabling efficient constrained optimization in real-time applications. For handling inequality constraints Eq. \ref{eq2}, we adopt a penalization strategy \cite{penopt}, where constraint elimination and optimality assurance follow the methodology outlined in \cite{c32}. This framework ultimately converts the original constrained formulation into an unconstrained optimization problem, and we use L-BFGS \cite{lbfgs} to solve it efficiently.
\section{Benchmarks and Experiments}
\label{sec:experiments}
We conduct extensive simulation and benchmark to demonstrate the effectiveness and advantages of our method in enabling efficient trajectory planning for large-scale formations and maintaining resilience in the presence of anomalous agents. The simulation platform is a PC with an Intel Core i7 8700K CPU running at 3.2 GHz and with 32-GB RAM at 3200 MHz. We empirically set the parameters of formation planning and RANSAC based on \cite{c1}, \cite{random}, which are present in Table \ref{para}.
\begin{table}[htbp]
\centering
\vspace{-0.2cm} 
\caption{Parameters for PCR-based formation planning}
\label{para} 
\begin{threeparttable}
\renewcommand{\arraystretch}{1.3} 
\setlength{\tabcolsep}{8pt} 
\begin{tabular}{lcr} 
\toprule
\textbf{Parameter} & \textbf{Symbol} & \textbf{Value} \\
\midrule
Inlier threshold of RANSAC      & $\tau_{ran}$          & 0.15 \\
Confidence of RANSAC            & $p_{ran}$             & 0.99 \\
Maximum iterations of RANSAC    & $k_{ran}$             & 1000 \\
Weight for formation similarity & $\mu_f$         & 300.0 \\
Weight for control effort       & $\mu_e$         & 80.0 \\
Weight for flight time          & $\mu_t$         & 80.0 \\
Weight for collision avoidance  & $\mu_c$         & 10000.0 \\
Weight for swarm reciprocal avoidance & $\mu_s$   & 10000.0 \\
Weight for dynamical feasibility& $\mu_d$         & 100.0 \\
\bottomrule
\end{tabular}
\end{threeparttable}
\vspace{-0.4cm} 
\end{table}

\subsection{Effective Large-Scale Formation Planning}
To study the computational efficiency and formation performance of our method, we analyze the local trajectory optimization time $t_{opt}$ and formation error $\bar e_{dist}$ (calculated as in \cite{zhou2025re}) with varying numbers of drones. We compare it with SOTA methods (i.e., Quan's \cite{c1}, Zhou's \cite{zhou2025re}).

\begin{figure}[!t]
    \centering
    \includegraphics[width=3.2in]{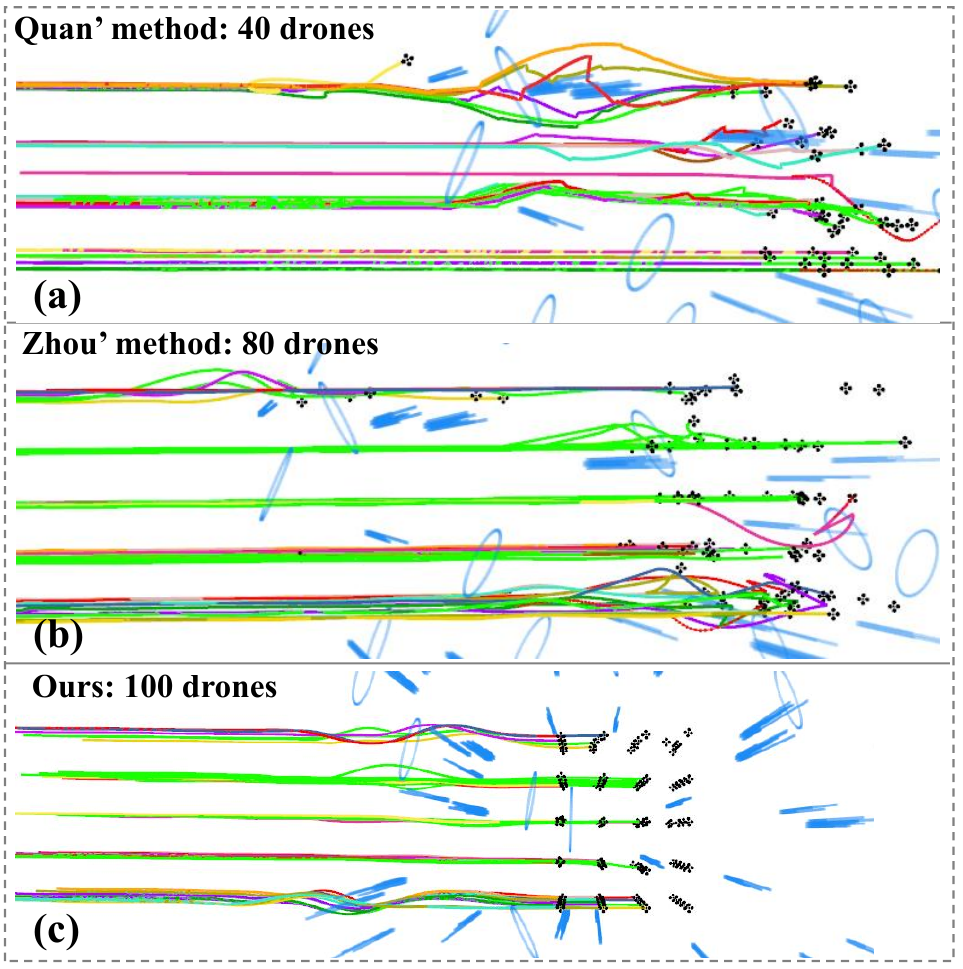}
    \vspace{-0.3cm}
    \caption{
    Experiments on computational efficiency and performance of formation planning across varying swarm number. (a)\&(b) Experiments on cubic formation maintenance using Quan's and Zhou's methods, respectively, both fail to sustain the formation effectively due to excessive trajectory communication latency induced by prolonged computation time. (c) Our method achieves high-performance cubic formation planning for 100 agents.
    }
    \vspace{-0.3cm}
    \label{Time_edist_compar2}
\end{figure}

As shown in Table \ref{tab1}, since Quan's uses the Laplacian operator which is $\mathcal O(N^2)$ complexity to measure the error with respect to the desired formation, $t_{\text{mean}}$ reaches 1.023s when the drone number is 40. As shown in Fig. \ref{Time_edist_compar2} (a), real-time trajectory planning can no longer be supported in this case, resulting in the failure of formation maintenance. Although Zhou's alleviates computation time by sparsing the complete graph, as the swarm number increases, it still fails to balance computation time and formation performance. We use recommended 30\% sparse connections in Zhou's here. Although Zhou's method achieves shorter planning time than Quan's, Fig. \ref{Time_edist_compar2} (b) shows that when the drone number reaches 100, $t_{\text{mean}}$ approaching 1s compromises formation maintenance.
In contrast, as shown in Table \ref{tab1}, our method achieves higher computational efficiency. Even when the drone number reaches 100, our trajectory computation time remains below 0.05 seconds and can ensure formation performance as shown in Fig. \ref{Time_edist_compar2} (c). Moreover, our trajectory optimization considers the positional information of all other agents, resulting in lower $\bar e_{dist}$. When the drone number is 100, the $\bar e_{dist} = 0.6065$ of ours is less than Zhou's. 

\begin{figure}[!t]
    \centering
    \includegraphics[width=3.1in]{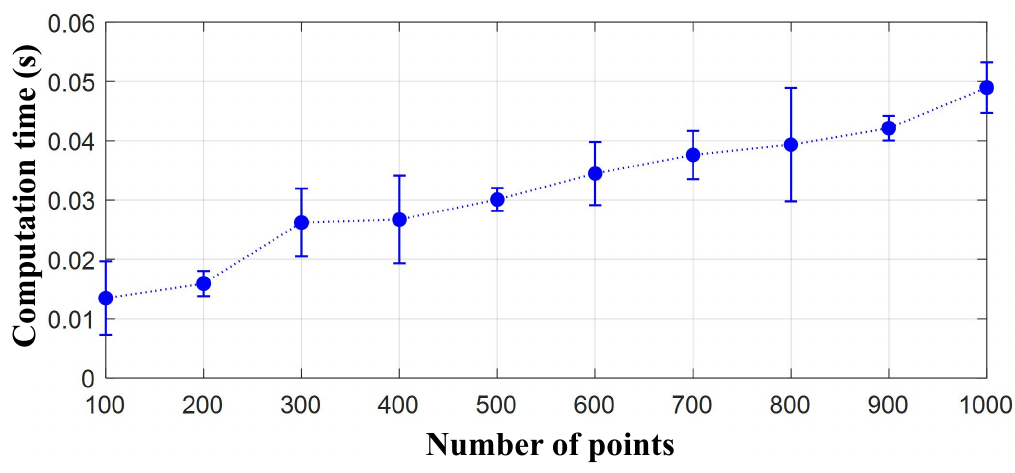}
    \vspace{-0.4cm}
    \caption{
    Line plot of mean and standard deviation of computation time versus point count (100 to 1000 points).
    }
    \vspace{-0.3cm}
    \label{1000}
    \vspace{-0.2cm}
\end{figure}

It is worth noting that in our method, the computational cost of formation trajectory planning does not primarily stem from OFPS computation. The increase in $t_{\text{mean}}$ is mainly attributed to two factors: (1) Each agent maintains a simulated local map, and the aggregation of numerous such maps consumes excessive computational resources. In practice, simulating maps for all agents on each robot is unnecessary, thereby real deployment incurs no such computational overhead. (2) collision avoidance checks among all agents. To demonstrate the extreme lightweight nature of our formation error computation (i.e., OFPS), we conduct an experiment that computes only the OFPS metric. In this experiment, we randomly generate desired formations varying number and then apply random perturbations to the desired formation positions to produce position sequences over a future horizon. As shown in Fig. \ref{1000}, the maximum OFPS computation time remains below 0.06s even with 1000 agents.
Finally, to further demonstrate the efficiency advantage of our method, as shown in Fig. \ref{fig1}, we deploy a formation planning experiment with 120 agents in the obstacle environment. Our method enables 120 agents to maintain a rocket-shaped formation, achieving an average per-agent local trajectory planning time of 0.09s and a $\bar e_{dist}$ of 0.6830.
\begin{table*}[thpb]
\centering 
\begin{threeparttable} 
\vspace{-0.5cm}
\renewcommand{\arraystretch}{1.2}
\setlength\tabcolsep{4.8pt} 
\caption{Performance comparison under varying drone counts}
\begin{tabular}{l *{15}{c}}
\toprule
\multirow{5}{*}{\textbf{Methods}} & \multicolumn{15}{c}{\textbf{Number of drones}} \\ 
\cmidrule(lr){2-16}
 & \multicolumn{3}{c}{\textbf{20}} & \multicolumn{3}{c}{\textbf{40}} & \multicolumn{3}{c}{\textbf{60}} & \multicolumn{3}{c}{\textbf{80}} & \multicolumn{3}{c}{\textbf{100}} \\
\cmidrule(lr){2-4}\cmidrule(lr){5-7}\cmidrule(lr){8-10}\cmidrule(lr){11-13}\cmidrule(lr){14-16}
 & $t_{\text{mean}}$ & $t_{\text{std}}$ & $\bar e_{\text{dist}}$ 
 & $t_{\text{mean}}$ & $t_{\text{std}}$ & $\bar e_{\text{dist}}$ 
 & $t_{\text{mean}}$ & $t_{\text{std}}$ & $\bar e_{\text{dist}}$ 
 & $t_{\text{mean}}$ & $t_{\text{std}}$ & $\bar e_{\text{dist}}$ 
 & $t_{\text{mean}}$ & $t_{\text{std}}$ & $\bar e_{\text{dist}}$ \\
\midrule
Quan's & 0.055 & 0.038 & \textbf{0.3403} & 1.023 & 0.765 & N/A & 1.848 & 1.015 & N/A & 3.053 & 1.700 & N/A & 5.814 & 4.312 & N/A \\
Zhou's & 0.025 & 0.017 & 0.4422 & 0.038 & 0.032 & 0.7620 & 0.073 & 0.082 & 0.9870 & 0.516 & 0.245 & 1.9631 & 0.933 & 0.561 & N/A \\
Ours   & \textbf{0.012} & \textbf{0.007} & 0.3680 & \textbf{0.014} & \textbf{0.007} & \textbf{0.6065} & \textbf{0.019} & \textbf{0.013} & \textbf{0.7355} & \textbf{0.026} & \textbf{0.018} & \textbf{0.9352} & \textbf{0.040} & \textbf{0.030} & \textbf{1.2580} \\
\bottomrule
\end{tabular}
\begin{tablenotes}
\item \footnotesize 
$t_{\text{mean}}$: mean planning computation time (s); $t_{\text{std}}$: standard deviation of planning time (s); $\bar e_{\text{dist}}$: formation distance error; N/A: planning failure (unable to achieve desired configuration in real time or $\bar e_{\text{dist}}>4.0$). 
\end{tablenotes}
\vspace{-0.5cm}
\label{tab1}
\end{threeparttable} 
\end{table*}

\begin{figure}[!t]
    \centering
    \includegraphics[width=3.5in]{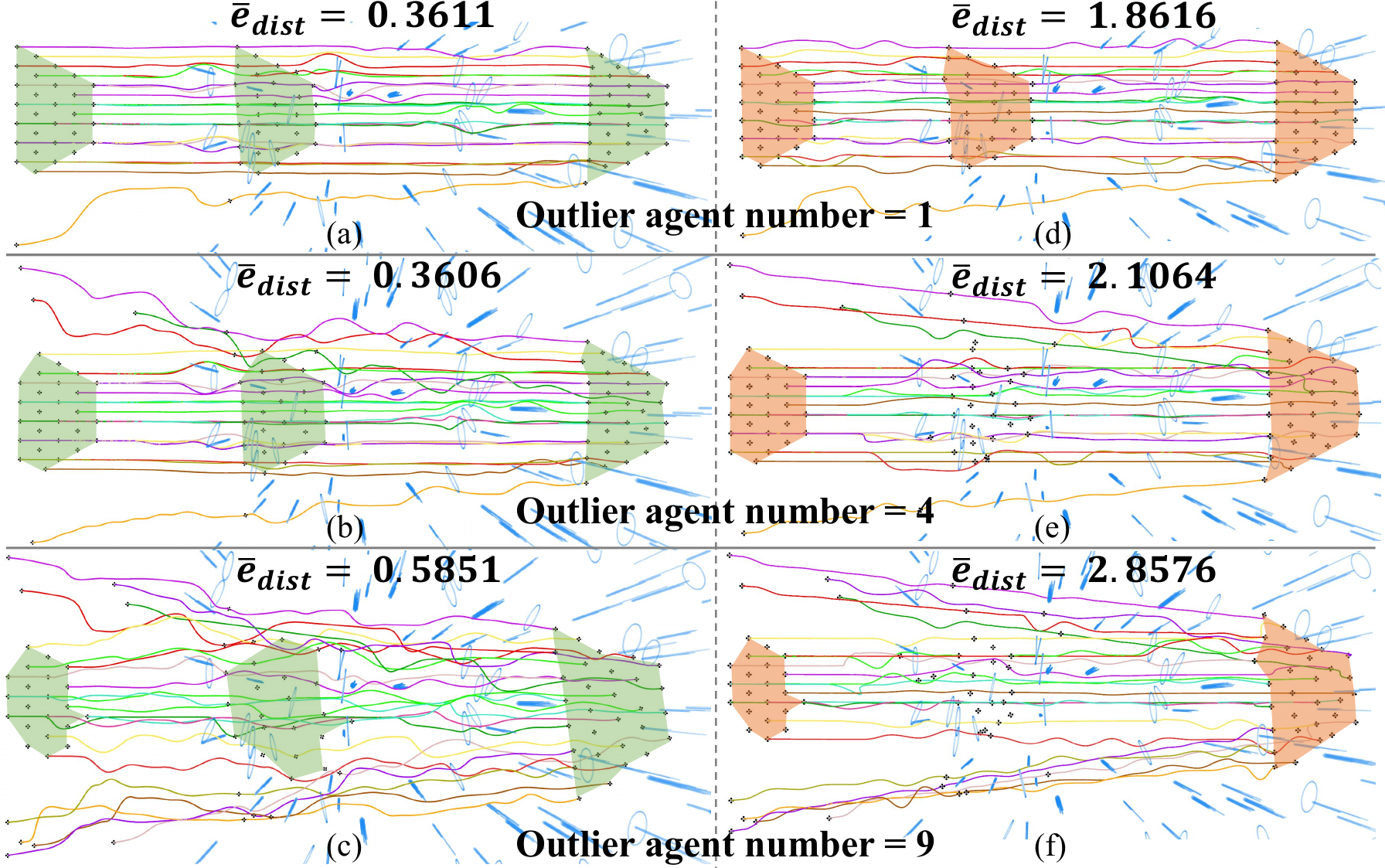}
    \vspace{-0.6cm}
    \caption{
    Formation planning with different numbers of outlier agents. (a)-(c) Our method performs in formation planning for 1, 3, and 9 outlier agents, respectively. (d)-(f) Zhou's method performs in formation planning for 1, 3, and 9 outlier agents, respectively.
    }
    \vspace{-0.3cm}
    \label{benchmark_ultra2}
    \vspace{-0.2cm}
\end{figure}

\subsection{Resilient Formation Planning}
To study the adaptive ability of our method to handle the abnormal agents and the locally optimal trajectories of other agents, we evaluate formation performance with varying numbers of the abnormal agents, and compare our method with other formation planning methods that adaptively handle abnormal agents.

As shown in Fig. \ref{benchmark_ultra2}, We test formation planning under three different proportions of abnormal outlier agents. As shown in Fig. \ref{benchmark_ultra2} (a)-(c), for Quan's method \cite{c1} without abnormal agent handling, the states of a small number of abnormal agents are propagated through the cooperative network, affecting all other agents and resulting in oscillations in the overall formation trajectories. Zhou's method \cite{zhou2025re} removes abnormal agents in advance by computing the maximum clique before trajectory planning. The coordination of the remaining agents is not affected for 3\% abnormal agents. However, when more abnormal agents are present, this single-step abnormal agent rejection method struggles to cope.

In contrast, our method eliminates abnormal agents during the computation of optimal spatiotemporal formation positions via the PCR algorithm with outlier rejection, meaning that each agent considers the abnormal states of others over a future time horizon along its entire trajectory. As shown in Fig. \ref{benchmark_ultra2} (c), even with 12\% abnormal agents, our method ensures performance, with $\bar e_{dist}$ of the remaining agents below 0.5. Furthermore, it is not necessary in practice for each agent to receive trajectories from all other agents. Even if some agents lose communication, we can still achieve registration between a subset of actual agents and all desired positions using PCR.

\subsection{Comparison under Various Formation Shape}
\begin{figure}[!t]
    \centering
    \includegraphics[width=3.0in]{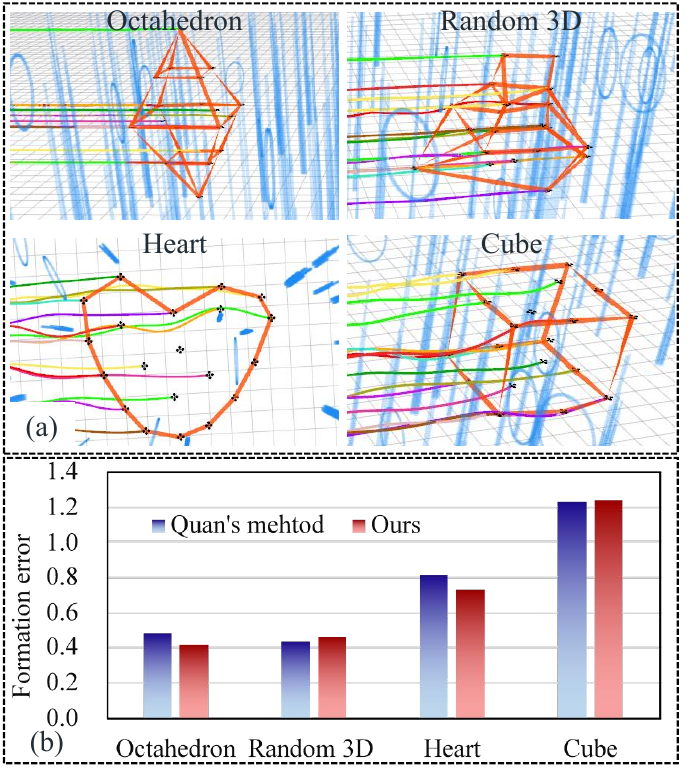}
    \vspace{-0.2cm}
    \caption{
    Comparative experiment of formation planning with different typical shapes. (a) Different typical formation shapes. (b) Comparison of formation errors $\bar e_{dist}$ between our method and Quan's method under different formation shapes.
    }
    \vspace{-0.4cm}
    \label{diff_shape}
\end{figure}

In this section, We further evaluate the performance of formations with various shapes and compare this performance against SOTA approaches for fully connected formation planning. To mitigate the adverse impact of prolonged trajectory optimization computation on distributed formation coordination, we consistently employ a fixed number of 20 agents to construct formations of different shapes, as illustrated in Fig. \ref{diff_shape} (a), including flat heart-shaped, cube, vertebral shape, and random three-dimensional shape. 
We test each shape 20 times in the obstacle environment and calculate the average formation error. The resulting formation error data are presented in Fig. \ref{diff_shape} (b). As evidenced across the four distinct formation shapes, the average formation error of our method remains consistently close to the SOTA method \cite{c1}. Since our method does not account for the influence of own current position on formation coordination when computing the optimal formation point, its performance is generally slightly inferior to Quan's method in most cases. Nevertheless, since our approach neglects only the contribution of a single agent, it still maintains strong formation performance.

\begin{figure}[!t]
    \centering
    \includegraphics[width=3.5in]{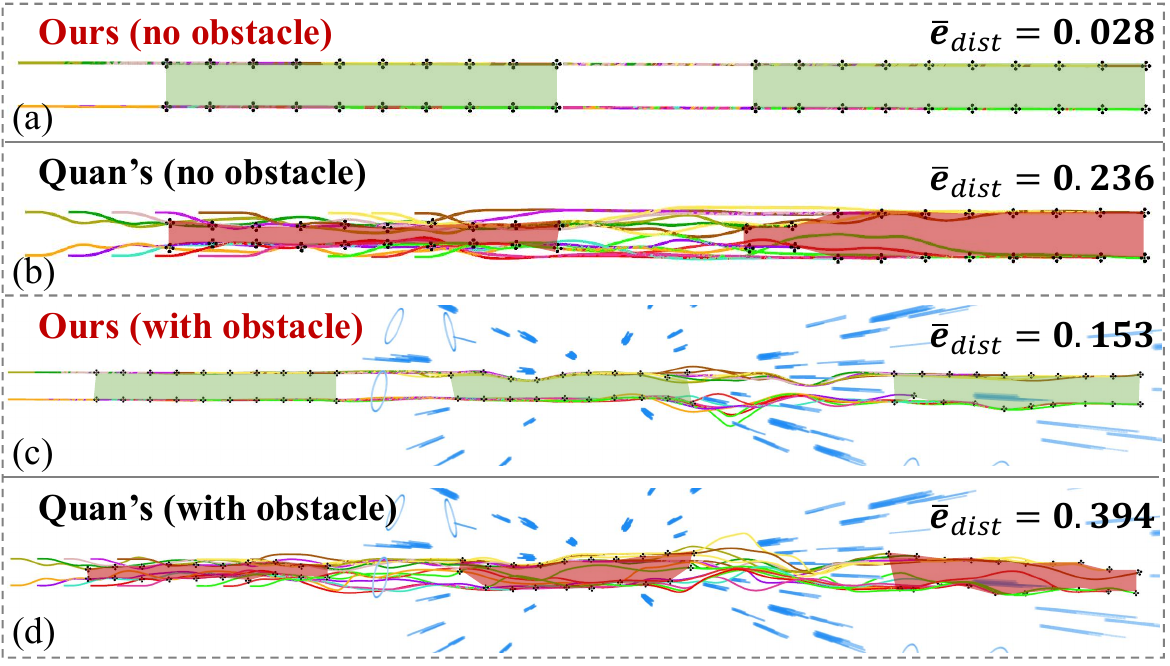}
    \vspace{-0.5cm}
    \caption{
    Comparative experiment of slender formation. (a)-(d) compares the motion trajectories and formation errors of our method and Quan's method while maintaining an elongated formation in obstacle free environment and obstacle environment. 
    }
    \vspace{-0.3cm}
    \label{coner}
    \vspace{-0.3cm}
\end{figure}

Furthermore, in comparison with methods that represent the desired formation using the graph Laplacian matrix \cite{c1}, \cite{zhou2025re}, our approach demonstrates superior formation performance in certain specific configurations. We evaluate an elongated rectangular formation that comprises 10 agents uniformly distributed along each long edge and 2 agents along each short edge, resulting in an overall dimension of 13.5 meters in length and 1.5 meters in width. We test formation planning for this elongated configuration in both obstacle free and obstacle environment.
We observe that in elongated formations characterized by the normalized Laplacian matrix, shape features along the long axis tend to dominate and obscure those along the short axis. As shown in Fig. \ref{coner}, this leads to reduced sensitivity to positional errors in the short-axis direction during motion, rendering the formation prone to deformation along the short axis and amplifying the overall formation error. Even without the disturbance of obstacles, due to perturbations arising from the optimizer's precision error, Quan's method is difficult to maintain a slender formation. In contrast, our method maintains consistent sensitivity to deviations in the short axis direction. Therefore, when there no obstacle, our $\bar e_{dist}$ is only $0.028$. And our formation error is smaller than the method based on normalized Laplacian matrix to characterize the formation in obstacles. In summary, our method can ensure satisfactory performance for large-scale formations and has better versatility for different formations.

\section{Conclusion and future work}
\label{sec:conclusion}

In this paper, we propose a large-scale distributed formation planning method based on PCR. We analyze the relationship between PCR and formation planning, and reformulate the problem of OFPS calculation as a series of spatiotemporal PCR problems. The PCR algorithm with outlier rejection is employed to obtain the OFPS, ensuring high efficiency and resilience in formation planning. Through simulation and benchmarks, we demonstrate that our algorithm can handle real-time, high-performance formation planning for swarms of over 100 agents. Even with nearly 10\% abnormal agents, our method ensures that the remaining normal agents can maintain formation.

In the future, we aim to realize even larger-scale formation planning by employing group-based strategies, using group-wise communication to reduce the dependence on high-bandwidth communication. Additionally, we plan to incorporate global path planning algorithms for formations to enhance the adaptability in more complex environments.

\section{Appendix}
\label{sec:appendix}
\emph{Proposition} 1: the points farther from the centroid of the point cloud have a greater influence on the point cloud registration outcome.

\textbf{\emph{Proof of \emph{Proposition}} 1:}  Consider a source point cloud $\mathbf P=\{ p_1, p_2, ..., p_{N_a} \}$ and a target point cloud $\mathbf Q = \{ q_1, q_2, ..., q_{N_a} \}$. The goal is to find a rotation matrix $\mathbf R$, a scale factor $s$, and a translation vector $\mathbf t$ that minimize the residual errors:
\begin{equation}
\begin{aligned}
E(\theta) = \sum_{i=1}^{N_a} \| q_i - (s \mathbf{R} p_i+\mathbf{t}) \|_2 ,
\end{aligned}
\end{equation}
the parameter vector is denoted as $\theta=[\mathbf R, \mathbf t, s]$. To simplify the analysis, we first centralize the point clouds by subtracting their centroids:
\begin{equation}
\begin{aligned}
\mu_{\mathbf P} = \frac{1}{N_a}\sum_{i=1}^{N_a}p_i,\ \ \ \  \mu_{\mathbf Q} = \frac{1}{N_a}\sum_{i=1}^{N_a}q_i, 
\end{aligned}
\end{equation}
\begin{equation}
\begin{aligned}
p'_i = p_i-\mu_p,\ \ \ \  q'_i = q_i-\mu_q.
\end{aligned}
\end{equation}
The residual function is $e_i =q_i - (s \mathbf{R} p_i+\mathbf{t})$. Its derivatives with respect to the parameters $\mathbf{R}$ and $s$ are:
\begin{equation}
\begin{aligned}
\frac{\partial e_i}{\partial \mathbf{R}} = -s \cdot p'_i\
,\quad 
\frac{\partial e_i}{\partial s} = -\mathbf{R} p'_i\
,\quad 
\frac{\partial e_i}{\partial \mathbf{t}} = - \mathbf{I}.
\end{aligned}
\end{equation}
The sensitivity of the residual to changes in $\mathbf{R}$ and $s$ is proportional to $p'_i\
$ (i.e. the point's distance from the centroid). And the Jacobian matrix $J_i$ for $e_i$ can be written as
\begin{equation}
\begin{aligned}
J_i = [s \cdot p'_i, -\mathbf{R} p'_i, - \mathbf{I}].
\end{aligned}
\end{equation}
The covariance matrix of the parameter estimates is given by the inverse of the Fisher Information Matrix (FIM):
\begin{equation}
\begin{aligned}
\text{Cov}(\hat{\theta}) \approx \sigma^2 (J^T J)^{-1},
\end{aligned}
\end{equation}
where $\hat{\theta}$ is the estimation of $\theta$, $J$ is the stack of all individual $J_i$ matrices. The information matrix is $J^T J = \sum_{i=1}^{N_a} J_i^T J_i$. Each point's contribution to this information matrix is $ H_i = J_i^T J_i $.
We analyze the terms related to point position in $H_i$:
\begin{equation}
\begin{aligned}
H_i(s,s) = \left( \frac{\partial e_i}{\partial s} \right)^T \left( \frac{\partial e_i}{\partial s} \right) = (R p_i')^T (R p_i') = p_i'^T p_i' = p_i' \
^2,
\end{aligned}
\end{equation}
\begin{equation}
\begin{aligned}
H_i(\mathbf{R},\mathbf{R}) = \left( \frac{\partial e_i}{\partial \mathbf{R}} \right)^T \left( \frac{\partial e_i}{\partial\mathbf{R}} \right) = s^2 ( p_i' )^T ( p_i' ) = s^2  p_i' \
^2,
\end{aligned}
\end{equation}
\begin{equation}
\begin{aligned}
H_i(t,t) = \left( \frac{\partial e_i}{\partial t} \right)^T \left( \frac{\partial e_i}{\partial t} \right) = I.
\end{aligned}
\end{equation}
Therefore, the parts of the information matrix $J^TJ$ related to scale and rotation depend on $ \sum_{i=1}^n p_i' \
^2$, while the part related to translation depends only on the number of points $n$. Since the covariance matrix is the inverse of the information matrix, a larger information contribution leads to lower uncertainty in the parameter estimates.
\hfill $\blacksquare$

\bibliography{ICRA2025zy}

@string{icra = {Proc. of the {IEEE} Intl. Conf. on Robot. and Autom.}}

@string{ram = {{IEEE} Robot. Autom. Mag.}}

@string{iros = {Proc. of the {IEEE/RSJ} Intl. Conf. on Intell. Robots and Syst.}}

@string{cdc = {Proc. of the {IEEE} Control and Decision Conf.}}

@article{c1,
  title={Robust and efficient trajectory planning for formation flight in dense environments},
  author={Quan, Lun and Yin, Longji and Zhang, Tingrui and Wang, Mingyang and Wang, Ruilin and Zhong, Sheng and Zhou, Xin and Cao, Yanjun and Xu, Chao and Gao, Fei},
  journal={IEEE Transactions on Robotics},
  year={2023}
}

@article{c15,
  title={Towards a swarm of agile micro quadrotors},
  author={Kushleyev, Alex and Mellinger, Daniel and Powers, Caitlin and Kumar, Vijay},
  journal={Autonomous Robots},
  volume={35},
  number={4},
  pages={287--300},
  year={2013},
  publisher={Springer}
}

@article{c16,
  title={Multi-robot formation control and object transport in dynamic environments via constrained optimization},
  author={Alonso-Mora, Javier and Baker, Stuart and Rus, Daniela},
  journal={The International Journal of Robotics Research},
  volume={36},
  number={9},
  pages={1000--1021},
  year={2017},
  publisher={SAGE Publications Sage UK: London, England}
}

@article{c18,
  title={Swarm of micro flying robots in the wild},
  author={Zhou, Xin and Wen, Xiangyong and Wang, Zhepei and Gao, Yuman and Li, Haojia and Wang, Qianhao and Yang, Tiankai and Lu, Haojian and Cao, Yanjun and Xu, Chao and others},
  journal={Science Robotics},
  volume={7},
  number={66},
  pages={eabm5954},
  year={2022},
  publisher={American Association for the Advancement of Science}
}

@article{c22,
  title={Finite-time formation control for multi-agent systems},
  author={Xiao, Feng and Wang, Long and Chen, Jie and Gao, Yanping},
  journal={Automatica},
  volume={45},
  number={11},
  pages={2605--2611},
  year={2009},
  publisher={Elsevier}
}

@article{c25,
  title={Region-based shape control for a swarm of robots},
  author={Cheah, Chien Chern and Hou, Saing Paul and Slotine, Jean Jacques E},
  journal={Automatica},
  volume={45},
  number={10},
  pages={2406--2411},
  year={2009},
  publisher={Elsevier}
}

@article{c32,
  title={Geometrically constrained trajectory optimization for multicopters},
  author={Wang, Zhepei and Zhou, Xin and Xu, Chao and Gao, Fei},
  journal={IEEE Transactions on Robotics},
  volume={38},
  number={5},
  pages={3259--3278},
  year={2022}
}

@article{1000agents,
  title={Programmable self-assembly in a thousand-robot swarm},
  author={Rubenstein, Michael and Cornejo, Alejandro and Nagpal, Radhika},
  journal={Science},
  volume={345},
  number={6198},
  pages={795--799},
  year={2014},
  publisher={American Association for the Advancement of Science}
}

@article{zhaoswarm,
  title={Mean-shift exploration in shape assembly of robot swarms},
  author={Sun, Guibin and Zhou, Rui and Ma, Zhao and Li, Yongqi and Gro{\ss}, Roderich and Chen, Zhang and Zhao, Shiyu},
  journal={Nature Communications},
  volume={14},
  number={1},
  pages={3476},
  year={2023},
  publisher={Nature Publishing Group UK London}
}

@article{1mswarm,
  title={Probabilistic and distributed control of a large-scale swarm of autonomous agents},
  author={Bandyopadhyay, Saptarshi and Chung, Soon-Jo and Hadaegh, Fred Y},
  journal={IEEE Transactions on Robotics},
  volume={33},
  number={5},
  pages={1103--1123},
  year={2017}
}

@article{resilient1,
  title={Resilient flocking for mobile robot teams},
  author={Saulnier, Kelsey and Saldana, David and Prorok, Amanda and Pappas, George J and Kumar, Vijay},
  journal={IEEE Robotics and Automation letters},
  volume={2},
  number={2},
  pages={1039--1046},
  year={2017}
}

@article{resilient2,
  title={Formations for resilient robot teams},
  author={Guerrero-Bonilla, Luis and Prorok, Amanda and Kumar, Vijay},
  journal={IEEE Robotics and Automation Letters},
  volume={2},
  number={2},
  pages={841--848},
  year={2017}
}

@inproceedings{maxcline,
  title={3D registration with maximal cliques},
  author={Zhang, Xiyu and Yang, Jiaqi and Zhang, Shikun and Zhang, Yanning},
  booktitle={Proceedings of the IEEE/CVF Conference on Computer Vision and Pattern Recognition},
  pages={17745--17754},
  year={2023}
}

@article{teaser,
  title={Teaser: Fast and certifiable point cloud registration},
  author={Yang, Heng and Shi, Jingnan and Carlone, Luca},
  journal={IEEE Transactions on Robotics},
  volume={37},
  number={2},
  pages={314--333},
  year={2020}
}

@inproceedings{pcm,
  title={Pairwise consistent measurement set maximization for robust multi-robot map merging},
  author={Mangelson, Joshua G and Dominic, Derrick and Eustice, Ryan M and Vasudevan, Ram},
  booktitle={2018 IEEE international conference on robotics and automation (ICRA)},
  pages={2916--2923}
}

@article{random,
  title={Random sample consensus: a paradigm for model fitting with applications to image analysis and automated cartography},
  author={Fischler, Martin A and Bolles, Robert C},
  journal={Communications of the ACM},
  volume={24},
  number={6},
  pages={381--395},
  year={1981},
  publisher={ACM New York, NY, USA}
}

@article{markov,
  title={Markov chain approach to probabilistic guidance for swarms of autonomous agents},
  author={A{\c{c}}{\i}kme{\c{s}}e, Beh{\c{c}}et and Bayard, David S},
  journal={Asian Journal of Control},
  volume={17},
  number={4},
  pages={1105--1124},
  year={2015},
  publisher={Wiley Online Library}
}

@article{vrb,
  title={Agile coordination and assistive collision avoidance for quadrotor swarms using virtual structures},
  author={Zhou, Dingjiang and Wang, Zijian and Schwager, Mac},
  journal={IEEE Transactions on Robotics},
  volume={34},
  number={4},
  pages={916--923},
  year={2018}
}

@inproceedings{peng,
  title={Obstacle avoidance of resilient UAV swarm formation with active sensing system in the dense environment},
  author={Peng, Peng and Dong, Wei and Chen, Gang and Zhu, Xiangyang},
  booktitle={2022 IEEE/RSJ International Conference on Intelligent Robots and Systems (IROS)},
  pages={10529--10535}
}

@article{learning2,
  title={Formation control with collision avoidance through deep reinforcement learning using model-guided demonstration},
  author={Sui, Zezhi and Pu, Zhiqiang and Yi, Jianqiang and Wu, Shiguang},
  journal={IEEE Transactions on Neural Networks and Learning Systems},
  volume={32},
  number={6},
  pages={2358--2372},
  year={2020}
}

@article{penopt,
  title={A computational algorithm for functional inequality constrained optimization problems},
  author={Jennings, Leslie Stephen and Teo, Kok Lay},
  journal={Automatica},
  volume={26},
  number={2},
  pages={371--375},
  year={1990},
  publisher={Elsevier}
}

@article{wuyi_formation,
  title={Multi-UAV Behavior-based Formation with Static and Dynamic Obstacles Avoidance via Reinforcement Learning},
  author={Xie, Yuqing and Yu, Chao and Zang, Hongzhi and Gao, Feng and Tang, Wenhao and Huang, Jingyi and Chen, Jiayu and Xu, Botian and Wu, Yi and Wang, Yu},
  journal={arXiv preprint arXiv:2410.18495},
  year={2024}
}

@inproceedings{relient3,
  title={Finite-time resilient formation control with bounded inputs},
  author={Usevitch, James and Garg, Kunal and Panagou, Dimitra},
  booktitle={2018 IEEE conference on decision and control (CDC)},
  pages={2567--2574}
}

@article{relient4,
  title={Distributed adaptive resilient formation control of uncertain nonholonomic mobile robots under deception attacks},
  author={Wang, Wei and Han, Zhen and Liu, Kexin and L{\"u}, Jinhu},
  journal={IEEE Transactions on Circuits and Systems I: Regular Papers},
  volume={68},
  number={9},
  pages={3822--3835},
  year={2021}
}

@article{zhou2025re,
  title={RE-Formation: Resilient and Efficient Formation Planning in Large-Scale Distributed Aerial Swarms},
  author={Zhou, Yuan and Quan, Lun and Xu, Chao and Xu, Guangtong and Gao, Fei},
  journal={IEEE Transactions on Automation Science and Engineering},
  year={2025}
}

@article{how_formation,
  title={A distributed pipeline for scalable, deconflicted formation flying},
  author={Lusk, Parker C and Cai, Xiaoyi and Wadhwania, Samir and Paris, Aleix and Fathian, Kaveh and How, Jonathan P},
  journal={IEEE Robotics and Automation Letters},
  volume={5},
  number={4},
  pages={5213--5220},
  year={2020}
}

@article{lbfgs,
  title={On the limited memory BFGS method for large scale optimization},
  author={Liu, Dong C and Nocedal, Jorge},
  journal={Mathematical programming},
  volume={45},
  number={1},
  pages={503--528},
  year={1989},
  publisher={Springer}
}
\end{document}